\documentclass[pdflatex,sn-mathphys-num]{sn-jnl}

\usepackage{graphicx}%
\usepackage{multirow}%
\usepackage{amsmath,amssymb,amsfonts}%
\usepackage{amsthm}%
\usepackage{mathrsfs}%
\usepackage[title]{appendix}%
\usepackage{xcolor}%
\usepackage{textcomp}%
\usepackage{manyfoot}%
\usepackage{booktabs}%
\usepackage{algorithm}%
\usepackage{algorithmicx}%
\usepackage{algpseudocode}%
\usepackage{listings}%
\usepackage{pdfpages}
\usepackage{dsfont}
\usepackage{subcaption}
\usepackage{pgffor}

\theoremstyle{thmstyleone}%
\theoremstyle{thmstyletwo}%

\theoremstyle{thmstylethree}%

\begin{document}

\title[Deep Learning–based Mechanical Characterization of Synthetically Generated Plant Fibers under Transverse Compression]{Deep Learning–based Mechanical Characterization of Synthetically Generated Plant Fibers under Transverse Compression}


\author*[1]{\fnm{Nikolaos} \sur{Kegkeroglou}}\email{nikolaos.kegkeroglou@thlabs.eu}

\author[2]{\fnm{Bilal} \sur{Sabillah}}\email{bilal.sabillah@femto-st.fr}

\author[3]{\fnm{Myrto} \sur{Inglezou}}\email{mi23878@essex.ac.uk}

\author[2]{\fnm{Vincent} \sur{Placet}}\email{vincent.placet@femto-st.fr}

\author[2]{\fnm{Cédric} \sur{Clevy}}\email{cedric.clevy@femto-st.fr}

\author[3]{\fnm{Antonios} \sur{Porichis}}\email{
ap21186@essex.ac.uk}

\affil*[1]{ \orgname{Tech Hive Labs}, \orgaddress{\street{Iliados 7}, \city{Halandri}, \postcode{152 32}, \country{Greece}}}

\affil[2]{\orgdiv{Université Marie et Louis Pasteur, SUPMICROTECH, CNRS}, \orgname{Institut FEMTO-ST}, \orgaddress{\city{Besançon}, \postcode{F-25000}, \country{France}}}

\affil[3]{\orgdiv{AI Innovation Centre}, \orgname{University of Essex}, \orgaddress{\street{Little Abington}, \city{Cambridge}, \postcode{CB21 6AL},  \country{UK}}}


\abstract{Plant fibers are increasingly used as sustainable reinforcements in composites, but the intrinsic variability of their microstructure causes substantial scatter in transverse mechanical behavior, complicating composite design. Characterizing this behavior currently relies on finite element (FE) simulations that require several minutes of high-performance computing per fiber, making high-throughput characterization across realistic fiber populations impractical. In this study, we propose a deep learning framework that replaces this FE bottleneck with a fast, accurate pipeline that includes two coupled recursive U-Nets that predict the step-by-step evolution of fiber deformation and internal stress fields under transverse compression, followed by a regression CNN that estimates the transverse Young's modulus from the final predicted stress field. Unlike prior deep learning approaches, which predict stress fields for a single fixed configuration or for idealized fiber geometries, the proposed framework recursively resolves the full temporal stress evolution for individual synthetic fibers derived from real hemp fiber cross-sections, including cases with and without a lumen. Trained and evaluated on 1,000 synthetically generated fiber geometries with FE-derived ground truth, the framework reproduces the stress field sequence with a Structural Similarity Index Measure (SSIM) of $0.9574 \pm 0.0074$ and a Root Mean Squared Error (RMSE) of $0.0345 \pm 0.0079$ and estimates the transverse Young's modulus with a $R^2$ coefficient of $0.938 \pm 0.024$, a Mean Absolute Error (MAE) of $24.31 \pm 1.23$ MPa and a Mean Absolute Percentage Error (MAPE) of $5.25 \pm 0.91\%$. Inference requires under one second per fiber, compared to approximately five minutes per FE simulation, representing an order-of-magnitude reduction in computational cost. This framework offers a fast and reliable tool for the mechanical characterization of individual synthetic plant fibers, with promising applications in the design of natural fiber-reinforced composites.}




\maketitle

\section{Introduction}\label{sec1}

Plant fibers are increasingly being employed as renewable reinforcement constituents in composite materials owing to their favorable specific mechanical properties and sustainability advantages \cite{doi:10.1177/204124791600700202}. Over the past decade, they have gained growing importance as reinforcing agents in a wide range of applications, including automotive lightweighting \cite{Skosana2024DrivingTS} and soft robotics \cite{https://doi.org/10.1002/aisy.202400414}. Extensive research efforts have focused on developing hybrid or fully biobased composites that combine plant fibers with biopolymer matrices, aiming to reduce reliance on non-renewable resources and mitigate environmental impact \cite{doi:10.1142/S0218625X25300035}.

Plant Fiber Reinforced Composites (PFRCs) have emerged as strong candidates to replace conventional synthetic fiber composites such as glass fiber–reinforced composites. It was projected that by 2025, bio-based fibers could account for up to 28\% of the total reinforcement materials market \cite{RAMESH2017558}. Compared to their synthetic counterparts, PFRCs exhibit competitive mechanical properties, while remaining environmentally friendly. Bast fibers such as flax, hemp, and nettle are considered high-performance fibers owing to their tensile strengths and favorable aspect ratios \cite{ARDANUY2015115}. These characteristics make them attractive candidates for both structural and functional bio-composite applications.

Nevertheless, a central obstacle to the wider adoption of PFRCs lies in the transverse mechanical behavior of the fibers themselves. In unidirectional composites, the transverse direction is typically the weakest, and local stress concentrations in this plane often govern the onset of damage. Characterizing the transverse stiffness and stress response of individual fibers is therefore essential, yet remains difficult. The Single Fiber Transverse Compression Test (SFTCT) has only recently been established as a direct means of probing this behavior in plant fibers \cite{GOVILAS2024108321}, and reliable measurements demand micro-mechatronic testing platforms with sub-micron precision combined with advanced microscopy \cite{PLACET2020S303}. Compounding this difficulty, plant fibers exhibit pronounced variability in microstructure and composition arising from differences in species, cultivation, harvesting, and processing, which translates into substantial scatter in mechanical performance and undermines predictable composite design \cite{doi:10.1177/0731684418756762}.

Extracting transverse properties from such tests further relies on inverse identification through finite element (FE) simulations. While these finite element analyses yield valuable mechanistic insight, they impose substantial computational costs that scale poorly with complex, highly variable microstructures. For instance, standard FE software such as Abaqus can require approximately 5 minutes to complete a single stress analysis \cite{jcs7080311}, and advanced frameworks modeling progressive composite deformation can require nearly 500 seconds per case \cite{CHEN2026100634}. This severe computational bottleneck renders high-throughput characterization across statistically representative fiber populations impractical. Consequently, this limitation motivates the search for fast, accurate surrogate models capable of bypassing repeated finite element evaluations. Deep learning architectures provide a highly efficient alternative. For example, specialized U-Net models have been shown to flawlessly predict complex local stress maps across diverse datasets with Structural Similarity Index Measure (SSIM) scores exceeding 0.97. Most notably, these surrogate frameworks reduce the required computational time from 5 minutes in traditional FE software to just 0.5 seconds on a standard laptop CPU, and down to a mere 0.05 seconds when utilizing GPU acceleration \cite{jcs7080311}. Implementing such a surrogate framework for plant fiber analysis offers the rapid, high-fidelity evaluation necessary to overcome traditional computational bottlenecks in bio-based composite characterization.


Deep learning has emerged as a powerful tool for solving conventional mechanics problems, including stress field analysis \cite{BHADURI2022109879} and mechanical property prediction \cite{Pathan2019}, and is increasingly applied to fiber-reinforced composites specifically \cite{Malashin2025}. These models learn patterns directly from large datasets, capturing the nonlinear relationships between geometry and mechanical response that are otherwise difficult to express in closed form. As a result, they achieve high accuracy and robustness across diverse microstructural configurations, while operating directly on image-based representations of the material that encode its full morphological information. Once trained, such models act as fast surrogates for the underlying physics, offering near-instantaneous evaluation across many samples or loading scenarios that would be prohibitively time-consuming with traditional finite element (FE) simulations \cite{jcs7080311}. This combination of accuracy and computational efficiency makes deep learning especially attractive for high-throughput tasks, such as screening large populations of fibers or exploring the design space of bio-based composites.

Recent advancements in materials science have seen a significant turn towards deep learning models for predicting the mechanical properties of composites \cite{D5TA00982K}. The typical approach involves training models on large datasets, often generated through FE methods, to capture the complex, non-linear relationships between a composite's microstructure and its mechanical performance. A common approach is to regress mechanical properties directly from key microstructural features. For example, Huang et al. \cite{HUANG2023116872} compared several regression techniques, including deep learning models to predict the effective elastic modulus based on the features of the constituent materials. Similarly, in \cite{Singh2025} various regression models were evaluated to predict the tensile, flexural, and impact strength of fiber composites using importance-selected features.

An alternative approach is to regress these properties directly from an image, which implicitly contains all the morphological information of the composite's phases. For instance, Gholami et al. \cite{jcs7020054} has used Convolutional Neural Networks (CNNs) to predict the Young's modulus and Poisson's ratio directly from microstructural images with a high degree of accuracy. A more recent work \cite{raaghav2025explainablepredictionmechanicalproperties} developed customized CNNs to predict the Young's modulus and yield strength of composites from 2D images. That work also employed explainability methods to verify that the models were learning scientifically relevant geometrical features, thereby increasing their trustworthiness for engineering applications. In a similar vein, Su et al. \cite{Su2023} utilized a deep learning method based on 3D CNN to predict the elastic coefficients of composite materials with inclusions of arbitrary sizes, shapes and material parameters. 

Beyond estimating material properties, deep learning is increasingly being applied to predict stress distributions. These models act as effective surrogate models that can bypass computationally expensive finite element simulations. Several studies have explored this approach for fiber-reinforced composites. In \cite{BHADURI2022109879}, a CNN based on the U-Net architecture \cite{ronneberger2015unetconvolutionalnetworksbiomedical} is used to predict stress field maps for composite material systems, consisting of circular fibers with varying spatial configurations. However, the use of idealized fiber shapes, such as perfect circles, may not capture the complexities of real-world materials. To address this limitation, \cite{jcs7080311} trained U-Net models on composite microstructures spanning uniform, non-uniform, and irregularly shaped fiber geometries to predict the corresponding von Mises stress maps. In \cite{jcs8100387}, they explore the application of StressNet \cite{Nie_2019}, a CNN originally proposed for linear elastic materials, on the prediction of stress field in 2D slices of segmented tomography images of a fiber-reinforced polymer specimen. In \cite{Jiang_2020}, a conditional generative adversarial network (cGAN) called StressGAN was proposed to predict stress distributions under varied geometry, load, and boundary conditions. Similarly, \cite{Yang_2021} utilized a cGAN architecture to map hierarchical composite microstructures directly to physical stress and strain fields across non-square tessellations like triangles and hexagons. In \cite{Gupta_2023}, a pix2pix cGAN model was developed to efficiently predict full-field principal stress maps based directly on microstructural images.

A smaller but growing body of work targets the temporal dimension of mechanical response, predicting how fields evolve under progressive loading rather than at a single instant. Wang et al. \cite{Wang2021} combined a Temporal Independent CNN with a Bidirectional Long Short-Term Memory (Bi-LSTM) network \cite{bilstm} to predict the sequence of maximum internal stress during fracture propagation in brittle materials, although it captures only a scalar quantity over time rather than the full spatial field. More recently, Chen et al. \cite{CHEN2026100634} introduced an auto-regressive U-Net that recursively predicts the coupled evolution of stress and damage fields throughout the deformation history of carbon fiber-reinforced composites, achieving order-of-magnitude speed-ups over interface-enriched finite element analysis.

While these works demonstrate the potential of deep learning for stress field prediction in composites, they predominantly predict a complete stress field for a single, fixed configuration rather than the step-by-step evolution of the complete spatial stress field as loading progresses. Moreover, they target idealized synthetic-fiber arrangements and have not addressed individual plant fibers, whose complex, realistic geometries and the potential presence of a lumen pose an additional challenge. Unlike the single-step or idealized-geometry approaches reviewed above, the present framework couples two recursive U-Nets to jointly resolve deformation and stress evolution across the full compression sequence for individual, non-idealized fiber geometries, a model construction not previously explored in this literature.

The aim of this study is to develop a deep learning framework capable of predicting both the sequential evolution of internal stress fields and the resulting transverse Young's modulus of individual plant fibers under transverse compression, at a fraction of the computational cost of finite element simulation. This aim is pursued through three specific objectives. The first objective is to generate a dataset of synthetic plant fiber geometries with FE-derived ground truth stress sequences, grounded in real hemp fiber morphology. The second objective is to design a recursive image-to-image architecture capable of resolving the temporal evolution of stress fields, a setting that, to our knowledge, has not previously been addressed for individual, non-idealized plant fiber geometries. The third objective is to couple this architecture with a regression model for direct estimation of the transverse Young's modulus from the predicted stress response. The novelty of this work lies primarily in the second objective. Prior deep learning approaches to composite stress field prediction, as discussed above, either predict a single static field or operate on idealized fiber arrangements. The present framework instead recursively predicts the full compression sequence for realistic, individually varying fiber cross sections.


\section{Methodology}\label{sec3}

This section describes the complete methodology underlying the proposed framework, from data generation to model design. We first detail the construction of the synthetic dataset, in which synthetic plant fiber cross-sections are generated and subjected to finite element simulations of transverse compression to obtain ground-truth stress field sequences, together with the apparent transverse Young's modulus recovered through inverse identification. We then describe the pre-processing operations that convert the raw simulation output into a representation suitable for deep learning. Finally, we present the modeling framework: a recursive, U-Net-based architecture that predicts the evolving deformation and stress fields step by step, coupled with a regression network that estimates the transverse Young's modulus from the final predicted stress field.

\subsection{Dataset Generation}

\begin{figure}[t]
    \centering
    \includegraphics[width=\textwidth]{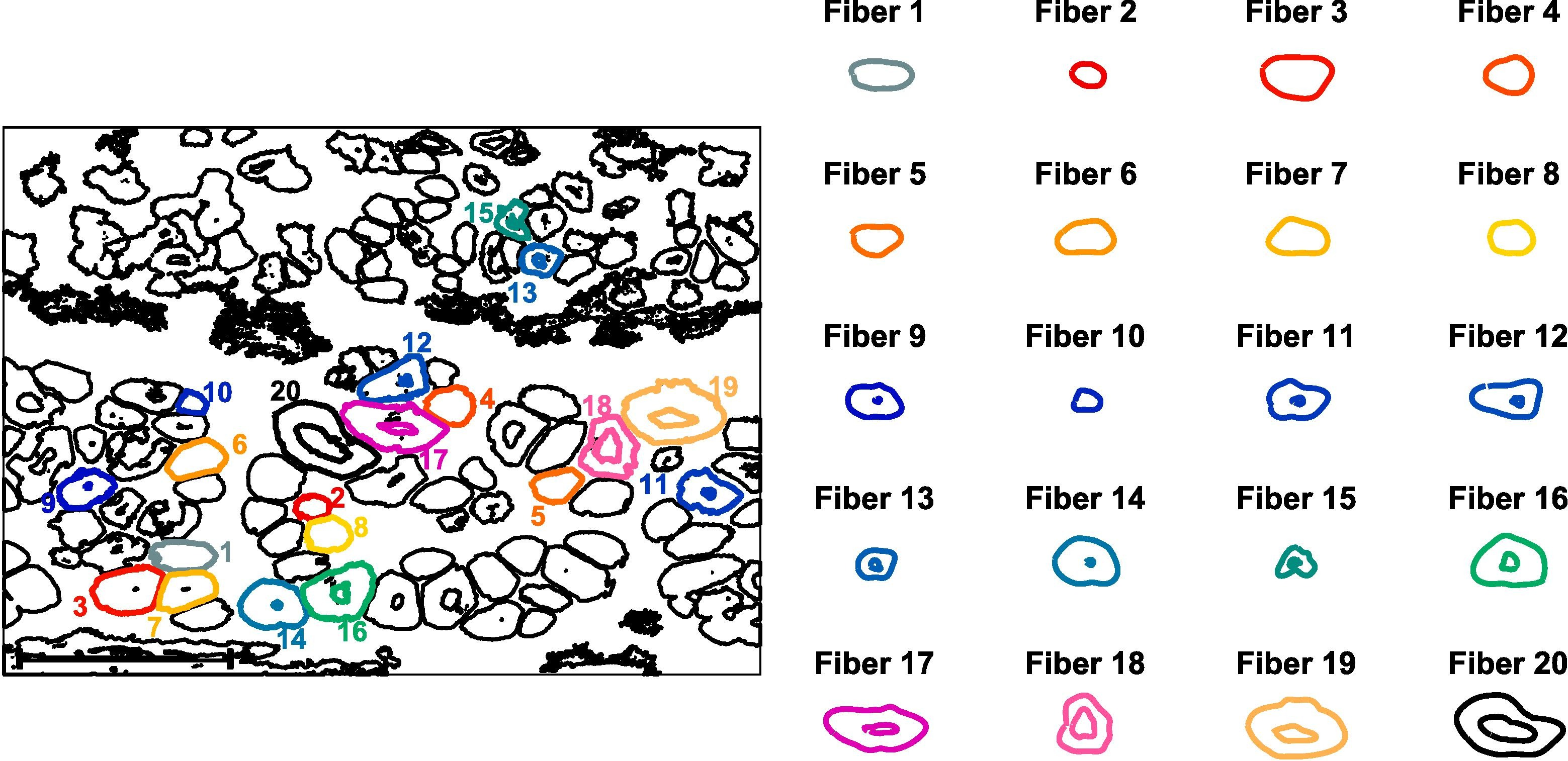}
    \caption{Extracted and smoothed hemp fiber cross-sectional geometries used for finite element analysis, showing examples of fibers with and without lumen. Reprinted from Composites Part A: Applied Science and Manufacturing, Vol 175, Jason Govilas, Cédric Clévy, Johnny Beaugrand, Vincent Placet, Investigating the influence of plant fiber geometry on apparent transverse elastic properties through finite element analysis, 107789, Copyright (2023), with permission from Elsevier \cite{GOVILAS2023107789}.}
    \label{fig:microscopy}
\end{figure}

The numerical workflow used to generate the synthetic fiber cross-sections is based on the method described in \cite{MATTRAND201410}. The generation process was based on the geometries of 20 real hemp fibers, whose cross-sectional contours were obtained from microscopic images of hemp fibers, comprising 9 fibers without lumen and 11 with lumen (Fig.~\ref{fig:microscopy}). Rather than treating these 20 fibers as a direct sample of the full population, the generation procedure treats them as a basis for interpolated morphological variation: the four geometric conformity filters (convexity, geodesic circularity, self-intersection, lumen boundary proximity) constrain the synthetic population to physically plausible cross-sections while the underlying generation method \cite{MATTRAND201410} introduces controlled variability in size, ellipticity, and lumen presence beyond what is directly observed in the source images. The resulting 1,000-fiber dataset, evenly split between lumen and non-lumen cases, is therefore better understood as a structured augmentation of the 20 base geometries than as 1,000 independent morphological samples. We acknowledge that this generation process inherits any systematic bias present in the original 20 fibers (e.g., their harvesting conditions or species-specific ellipticity), and note that validating the framework against fibers derived from an independently sourced set of cross-sections is an important direction for future work.

\begin{figure}[t!]
    \centering
    \includegraphics[width=\textwidth]{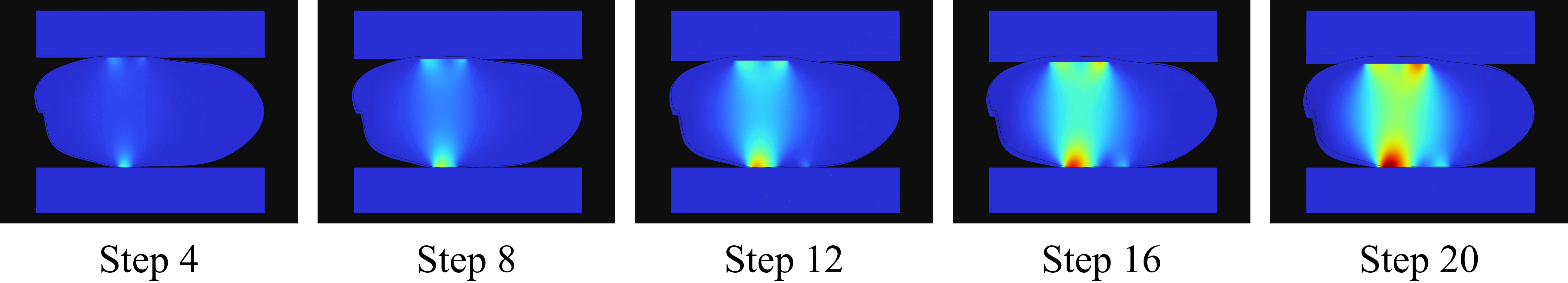}
    \caption{The progressive evolution of the internal stress field output from the FE transverse compression simulations at selected loading intervals.}
    \label{fig:fem_steps}
\end{figure}

Two-dimensional plane strain finite element simulations of transverse compression were conducted using COMSOL Multiphysics® (version 6.2) to resolve the internal stress distributions. The numerical framework, including the boundary conditions and contact mechanics, replicates the model validated in \cite{govilas:tel-04472025}. The simulation computes the stress fields at successive steps under the applied load, producing a sequence of images that shows the evolution of stress throughout the material (Fig.~\ref{fig:fem_steps}). The compression is driven by a rigid upper platen over 20 discrete displacement steps up to a maximum total displacement of $1~\mathrm{\mu m}$ (yielding an incremental step size of $0.05~\mathrm{\mu m}$ per step). The fiber material is modeled as linear elastic and transversely isotropic, defined by the following parameters:
\[
E_\mathrm{L} = 50~\mathrm{GPa}, \quad E_\mathrm{T} = 1~\mathrm{GPa}, \quad \nu_\mathrm{TT} = 0.07, \quad \nu_\mathrm{LT} = 0, \quad G_\mathrm{TT} = \frac{E_\mathrm{T}}{2(1+\nu_\mathrm{TT})}
\] 
where $E_\mathrm{L}$ and $E_\mathrm{T}$ represent the longitudinal and transverse Young's moduli, $\nu_\mathrm{LT}$ and $\nu_\mathrm{TT}$ are the Poisson's ratios in their respective planes, and $G_\mathrm{TT}$ is the transverse shear modulus. This linear elastic, transversely isotropic idealization is a reasonable approximation for the small-strain regime considered here (maximum platen displacement of 1 µm), though it does not capture the nonlinear or viscoelastic responses plant fibers can exhibit at larger deformations or under sustained loading. Extending the framework to such behavior would require nonlinear or viscoelastic constitutive models in the FE simulations and corresponding architectural changes to the network. The underlying nonlinear equilibrium equations were resolved via the Multifrontal Massively Parallel Sparse Direct Solver (MUMPS). All stationary computations were performed on a high-performance computing cluster (512~GB RAM, 2.5~GHz processor architecture), requiring an average computation time of approximately 5 minutes per fiber.

Following the forward simulations, the apparent transverse Young's modulus ($E_{\mathrm{T,app}}$) of each fiber was determined via an inverse identification procedure based on experimental protocols. The compression force per unit length ($F_\mathrm{L}$) was evaluated at each step by integrating the vertical component of the contact pressure across the contact boundary between the fiber outer boundary and the upper platen. The total radial deformation ($U$) corresponds directly to the prescribed platen displacement. These variables, alongside the geometric characteristics of the fiber cross-sections, were used to populate Jawad's analytical formulation~\cite{jawad1978} for the Single Fiber Transverse Compression Test (SFTCT):
\begin{equation}
U = \frac{4 F_\mathrm{L}}{\pi}\left[ S_{11}\left(\sinh^{-1}\left(\tfrac{r}{b}\right) + \ln(2) \right) - \tfrac{1}{2}\left(S_{11} + S_{12}\right) + S_{12} \frac{R}{b}\sqrt{1 + \left(\tfrac{r}{b}\right)^2} - \tfrac{r}{b} \right]
\end{equation}
where $b$ is the contact semi-width, $R$ and $r$ denote the major and minor radii of an ellipse fitted to the outer fiber contour, and $S_{11}=1/E_{\mathrm{T,app}}$ and $S_{12}=-(\nu_\mathrm{TT}/E_{\mathrm{T,app}})$ represent the components of the compliance matrix. A least-squares regression optimization was implemented to fit this analytical expression to the force-displacement data generated by the finite element model, enabling the unique identification of $E_{\mathrm{T,app}}$ for each sample.

Jawad's analytical formulation was originally derived under the assumption of a circular fiber cross-section under transverse compression. In the present framework, an ellipse is fitted to each synthetic fiber's outer contour (radii R and r) to approximate this circular assumption for non-circular geometries. This approximation introduces some error in the inverse-identified $E_{\mathrm{T,app}}$ for fibers with pronounced irregularity or high aspect ratio, since the true stress distribution around a non-elliptical contact boundary deviates from the idealized case the formulation assumes. Because $E_{\mathrm{T,app}}$ serves as the ground-truth target for the regression model, this approximation error propagates into the reported $R^2$, MAE, and MAPE values. We note this as a source of label noise inherent to the inverse identification protocol rather than a limitation of the deep learning framework itself.
\subsection{Pre-processing}

\begin{figure}[t]
    \centering
    \includegraphics[width=\textwidth]{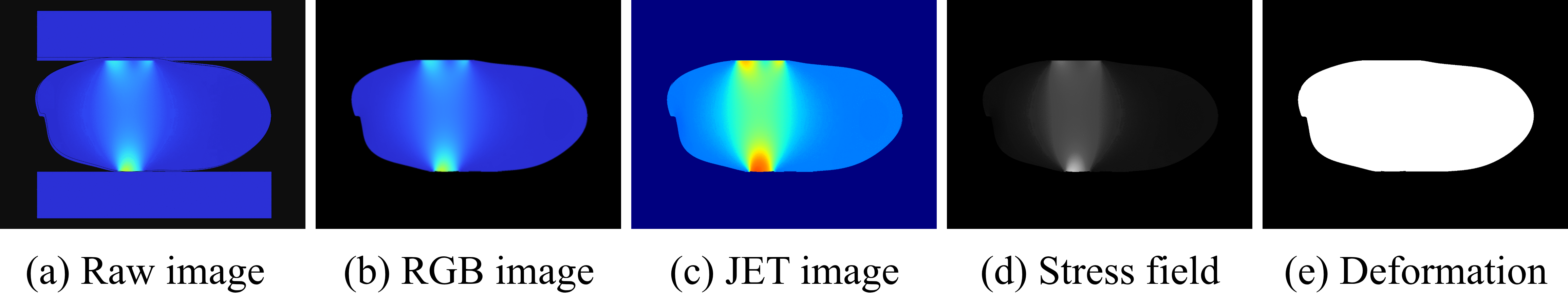}
    \caption{The pre-processing steps applied to the FE output sequences.}
    \label{fig:modalities}
\end{figure}

Prior to model training, we apply a series of pre-processing operations to the generated dataset to ensure its suitability for deep learning. Fig.~\ref{fig:modalities} illustrates an example applied on step 8 of the FE output sequence. The raw compression image sequence from the simulation contains several undesired visual elements, including the upper and lower platens of the simulation setup and a black contour surrounding the fiber specimen (Fig.~\ref{fig:modalities} (a)). These components do not contribute to useful visual information and instead introduce noise that can hinder the learning process.

To address this, basic image processing techniques are employed to automatically remove these regions, i.e., boundary-based masking for the platens and average pooling for the black contour pixels, resulting in a cleaner and more representative dataset for subsequent model training (Fig.~\ref{fig:modalities} (b)). Next, the resulting stress fields are visualized using the JET colormap, where stress increases from blue to red (Fig.~\ref{fig:modalities} (c)). To convert these images to grayscale while preserving the relative ordering of stress values within each sample, a per-sample normalisation strategy is adopted. For each sample, all distinct RGB values from the JET colormap appearing across the entire stress field sequence are first extracted and ordered according to their position in the colormap, from blue (low stress) to red (high stress), yielding an ordered set $D$. The grayscale intensity range [0,1] is then discretized into 256 equally spaced bins. Each pixel is assigned a grayscale value based on the index of its RGB value within $D$, so that lower-stress (more blue) pixels map to darker intensities and higher-stress (more red) pixels map to brighter intensities. This procedure yields grayscale images that represent the normalized stress field throughout the compression sequence (Fig.~\ref{fig:modalities} (d)). Following grayscale conversion, the normalized stress fields are scaled to physical units using a global min-max scaling procedure. The minimum value is defined as 0, and the maximum value is derived from the peak stress observed across the entire simulation dataset. This preserves the physical scale of the stress values within a normalized $[0,1]$ range optimized for model training. Lastly, binary masks of the fiber are extracted from the non-zero pixels in each grayscale frame, capturing the deformation of the fiber throughout the compression sequence (Fig.~\ref{fig:modalities} (e)).

Although raw scalar stress matrices are available directly from the finite element solver, the visual JET-colormap export was used as the basis for this conversion rather than the scalar field itself, since the FE solver computes stress over an unstructured, adaptively refined mesh, whereas the U-Net architecture requires a structured, uniform cartesian grid. The colormap image export inherently provides this uniform grid while remaining perfectly spatially aligned, pixel for pixel, with the binary deformation masks extracted from the same rendered frames. Mapping the unstructured scalar mesh data onto a uniform grid directly would instead require an additional spatial interpolation step, with a risk of introducing its own artifacts near the highly-stressed contact boundaries. This choice, however, subjects the underlying continuous stress field to uniform quantization, since the 256 discretized grayscale bins impose a maximum theoretical error on any given pixel of half the step size between bins:
\begin{equation}
e_{max} = \pm \frac{S_{max}}{2 \times 255} \approx \pm 0.00196 \cdot S_{max}
\end{equation}
which bounds the maximum information loss introduced by the colormap conversion to under $0.2\%$ of the global peak stress $S_{max}$.


The decision to scale the stress fields using a dataset-wide maximum, rather than normalizing each fiber independently to its own $[0,1]$ range, was similarly driven by the requirements of the downstream regression task. An alternative architecture was also evaluated, in which the Stress U-Net was instead trained on per-sample normalized targets, and a separate parallel CNN, taking only the undeformed binary mask as input, was tasked with predicting each fiber's peak stress to rescale the U-Net output back into physical units. While the per-sample-normalized Stress U-Net achieved strong structural fidelity in this configuration, the parallel CNN could not reliably infer peak stress from the undeformed mask alone, and the errors compounded through the rescaling step degraded the resulting physical stress reconstructions. The adopted dataset-wide scaling instead predicts physical stress fields directly, without requiring this auxiliary rescaling step, which in turn serve as the direct input to the Modulus CNN. Under this scheme, fibers with low peak stress occupy a narrower band of the $[0,1]$ range than high-stress fibers. However, this does not measurably limit the spatial fidelity of predictions on these samples, since (i) training is performed at 32-bit floating-point precision, providing ample granularity to resolve spatial stress gradients even within a compressed range, and (ii) the RMSE metric used to evaluate spatial fidelity (Section 4.2) rescales each predicted and ground-truth field to its own $[0,1]$ range prior to error computation, decoupling the evaluation of spatial distribution accuracy from absolute magnitude. 


Let the sequence of physical stress field images be $\mathbf{S} = \{ S_i \}_{i=0}^{N-1}$ and the sequence of fiber deformation masks be $\mathbf{B} = \{ B_i \}_{i=0}^{N-1}$, where $N$ is the number of steps in each compression sequence. These image sequences, $\mathbf{S}$ and $\mathbf{B}$, are used for self-supervised training of an image-to-image model, which is described in detail later in this work. To further increase the diversity of the training data, online data augmentation is applied via a stochastic spatial transformation operator, $\mathcal{T}_{\text{flip}}$. During training, horizontal flips are applied with a 50\% probability such that $\mathbf{S}_{\text{aug}} = \mathcal{T}_{\text{flip}}(\mathbf{S})$ and $\mathbf{B}_{\text{aug}} = \mathcal{T}_{\text{flip}}(\mathbf{B})$. Crucially, the operator is applied uniformly across all $N$ frame steps within a given sequence to preserve the exact spatial and temporal continuity between the fiber masks and their corresponding evolving stress fields. Horizontal flipping was selected as the sole augmentation strategy because it is the only spatial transformation that preserves the physical validity of the simulation setup without modification. Rotations would misalign the fiber's major and minor axes with the platen's fixed vertical loading direction, and would therefore require re-simulating the compression response rather than simply transforming existing images. Scaling transformations would alter the fiber's apparent size without a corresponding change to its mechanical response, since physical size is already separately encoded through the normalized radii (R, r). Elastic deformations could introduce fiber shapes inconsistent with the geometric conformity filters applied during dataset generation (Section 3.1), potentially producing samples with no valid corresponding stress response.

In addition, since the simulation-generated images are stored at a fixed pixel resolution that does not encode the physical dimensions of the fibers, fibers with substantially different true sizes may appear visually similar in the image domain. As the physical dimensions strongly influence the mechanical response of the fiber under compression, they constitute essential information that must be explicitly incorporated into the prediction pipeline. To this end, the major and minor radii ($R$, $r$) of each fiber sample are normalized using a global min-max scaling procedure similar to the one used for the stress fields, mapping them to a $[0,1]$ range based on the dataset-wide minimum and maximum values. These normalized scalars are retained as part of the sample representation and are later fed directly into the image-to-image model, as described in the following section.

Alongside this, a regression model is trained to predict the mechanical properties, specifically the transverse Young’s modulus ($E_{\mathrm{T}}$), as derived from the analytical model \cite{jawad1978}. In the proposed unified framework, the network takes as input the final predicted stress field generated by the stress image-to-image model. This formulation allows the learned representation of the full compression response to directly inform the estimation of effective material properties.

\subsection{Modeling Framework}
\begin{figure}[t!]
    \centering
    \includegraphics[width=\textwidth]{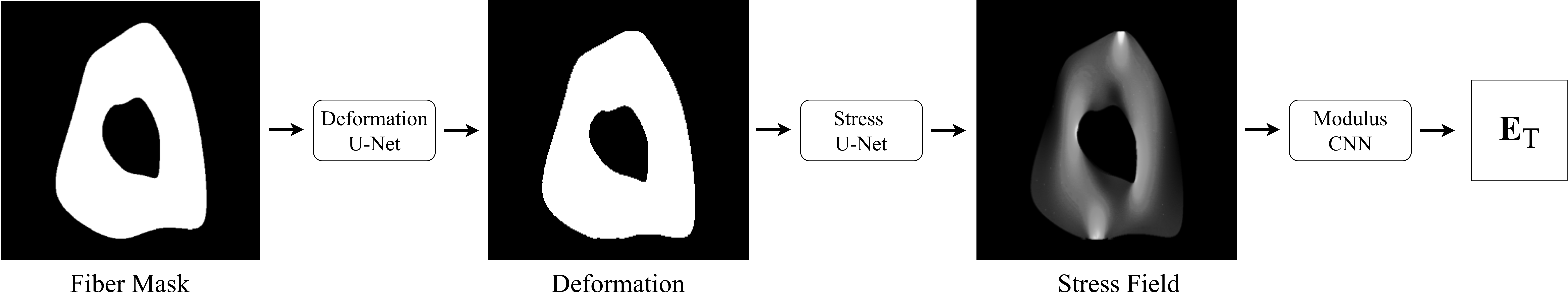}
    \caption{Schematic representation of the deep learning pipeline. Starting from the initial fiber mask, the Deformation U-Net predicts the evolving fiber geometry step by step. At each step, the Stress U-Net predicts the corresponding physical stress field, conditioned on the predicted deformation. The final predicted stress field is then passed to the Modulus CNN, which estimates the transverse Young's modulus ($E_T$).}
    \label{fig:architecture}
\end{figure}

The overall modeling framework consists of three components, each specialized for a distinct aspect of the fiber mechanics, which together enable accurate prediction of both the evolving stress field and the resulting mechanical properties. Fig.~\ref{fig:architecture} illustrates how the three models work together: the Deformation U-Net predicts the fiber deformation sequence step by step from the initial fiber mask and the Stress U-Net predicts the physical stress field sequence over the same steps, while the Modulus CNN predicts the transverse Young's modulus from the final predicted stress field.

\subsubsection*{Image-to-Image Model}

To model the coupled evolution of fiber deformation and internal stress fields across the compression sequence, we utilize a recursive U-Net-based image-to-image model. Its architecture follows the standard U-Net formulation, consisting of an encoder–decoder structure with skip connections. The encoder is composed of convolutional blocks with batch normalization and ReLU activation, interleaved with max-pooling to reduce spatial resolution, while the decoder symmetrically upsamples the feature maps using transposed convolutions and concatenates them with encoder features to preserve spatial detail. A final \(1 \times 1\) convolution followed by a sigmoid activation produces a pixel-wise output image. 

Two instances of this model are trained. The first instance, referred to as the Deformation U-Net ($f$), is used for deformation prediction. It takes as input the fiber mask \(B_i\) at time step \(i\), and learns to predict the future deformation state \(B_{i+\Delta}\), where \(\Delta\) is a fixed stride in the compression sequence. The second instance, referred to as the Stress U-Net ($g$), is used for stress prediction. It takes as input the triplet \((S_i, B_i, \hat{B}_{i+\Delta})\), consisting of the current stress field, the current fiber mask, and the predicted next-step mask from $f$, together with the normalized ellipse axis lengths. It learns to predict the corresponding future physical stress field \(S_{i+\Delta}\). Formally, the recursive prediction process is expressed as:
\[
\hat{B}_{i+\Delta} = f_{\theta}(B_i), 
\qquad
\hat{S}_{i+\Delta} = g_{\phi}(S_i, B_i, \hat{B}_{i+\Delta})
\]
where \(\theta\) and \(\phi\), denote the learnable parameters of the two network instances $f$ and $g$, respectively. The normalized radii ($R$, $r$) are incorporated into $g$ by concatenating them with the latent embedding produced by the bottleneck fully connected layer of the encoder. The resulting ($d$+2)-dimensional feature vector, where $d$ denotes the bottleneck dimension, is then provided as input to the corresponding fully connected layer in the decoder before feature reconstruction, enabling $g$ to distinguish between fibers that share similar image representations but exhibit different stress responses due to their physical dimensions.

Both networks are trained in a self-supervised manner using the simulation-derived deformation and stress sequences \(\mathbf{B} = \{B_i\}_{i=0}^{N-1}\) and \(\mathbf{S} = \{S_i\}_{i=0}^{N-1}\). For $f$, the training objective is the pixel-wise binary cross-entropy (BCE) loss between the predicted and ground-truth deformation masks:
\begin{equation}
\mathcal{L}_{\text{def}} = 
- \frac{1}{M (N - \Delta + 1)} \sum_{j=1}^{M} \sum_{i=0}^{N-\Delta} \sum_{p} 
\Big[ B_{i+\Delta}^{j}(p) \log \hat{B}_{i+\Delta}^{j}(p) + 
(1 - B_{i+\Delta}^{j}(p)) \log (1 - \hat{B}_{i+\Delta}^{j}(p)) \Big]
\end{equation}
where $M$ is the number of fiber samples and $p$ indexes pixels. For $g$, training is performed by minimizing the Mean Squared Error (MSE) between predicted and ground-truth stress fields at the corresponding future time step:
\begin{equation}
\mathcal{L}_{\text{str}} = 
\frac{1}{M (N - \Delta + 1)} \sum_{j=1}^{M} \sum_{i=0}^{N-\Delta} 
\big\| \hat{S}_{i+\Delta}^{j} - S_{i+\Delta}^{j} \big\|_2^2
\end{equation}

It is important to note that during training of $g$, the predicted masks $\hat{B}_{i+\Delta}$ from $f$ are used as inputs rather than the ground-truth FE-derived masks $B_{i+\Delta}$. This ensures that the input distribution of $g$ at training time matches the distribution it will encounter during inference, where only predicted future masks are available.

\subsubsection*{Regression Model}

To predict the transverse Young's modulus $E_{\mathrm{T}}$, we train a regression model, referred to as the Modulus CNN ($h$), which consists of a sequence of convolutional blocks, each containing two convolutional layers followed by batch normalization and ReLU activation. Max-pooling is applied after each block to progressively reduce spatial resolution, while the number of filters doubles across layers, forming a pyramidal encoder. The output of the final convolutional block is flattened into a bottleneck representation and passed through a hidden and an output fully connected layer to produce the mechanical property prediction. Model training is performed in a supervised setting using the final predicted stress field \(S_{N-1}^{j}\) as input, the apparent transverse Young's modulus $E_{\mathrm{T,app}}$ as ground-truth target, and is guided by a Mean Squared Error (MSE) loss:
\begin{equation}
\mathcal{L}_{\text{reg}} = \frac{1}{M}\sum_{j=1}^{M} \left( \hat{E}_{\mathrm{T}}^{j} - E_{\mathrm{T,app}}^{j} \right)^{2}
\end{equation}
where $\hat{E_{\mathrm{T}}^{j}}$ denotes the output prediction.

\section{Results}\label{sec4}

We evaluate the proposed framework along two complementary axes that mirror its predictive tasks: the fidelity of the predicted stress field sequences and the accuracy of the estimated transverse Young's modulus. In both cases, the framework is benchmarked against representative baseline architectures drawn from the stress field and mechanical property prediction literature, using quantitative metrics alongside qualitative comparisons against the finite element ground truth. To ensure that the observed performance is not an artifact of a particular initialization, all experiments are repeated across three random seeds and reported as means with standard deviations. Beyond predictive accuracy, we assess the computational efficiency of the framework relative to the finite element simulations it is designed to replace, and conduct an ablation study to isolate the individual contributions of the recursive formulation, the deformation branch, and the choice of stride. We note that, beyond the structural fidelity metrics (SSIM, RMSE) reported for the predicted stress fields, the accuracy of the downstream Modulus CNN provides a functional, physics-based validation of these fields: since the transverse Young's modulus is estimated solely from the predicted stress field, an accurate modulus prediction is only possible if the underlying stress field correctly captures the spatial distribution and magnitude of the internal stress concentrations.

\subsection{Implementation Details}

\begin{table}[t!]
\centering
\caption{Hyperparameters of the deep learning models}
\begin{tabular}{l l l}
\hline
\textbf{Hyperparameter} & \textbf{Regression} & \textbf{Image-to-Image} \\
\hline
Input size & $(256, 256)$ & $(256, 256)$ \\
Number of layers & 4 & 4 \\
Number of initial filters & 64 & 32 \\
Convolution kernel & $3 \times 3$ & $3 \times 3$ \\
Pooling size & 4 & 2 \\
Bottleneck dimension & 32 & 128 \\
Hidden dimension & 16 & - \\
Output dimension & 1 & $(256, 256)$ \\
\hline
\end{tabular}
\label{tab:hyperparams}
\end{table}

\begin{figure}[t]
    \centering
    \begin{subfigure}[b]{0.45\textwidth}
        \includegraphics[width=\textwidth]{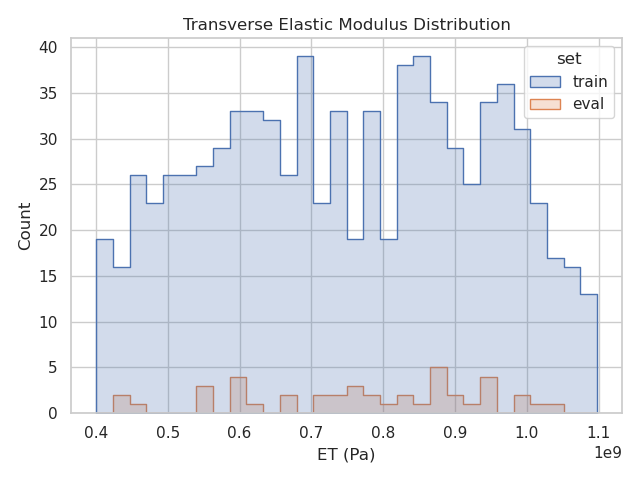}
        \caption{Seed 1}
    \end{subfigure}
    \begin{subfigure}[b]{0.45\textwidth}
        \includegraphics[width=\textwidth]{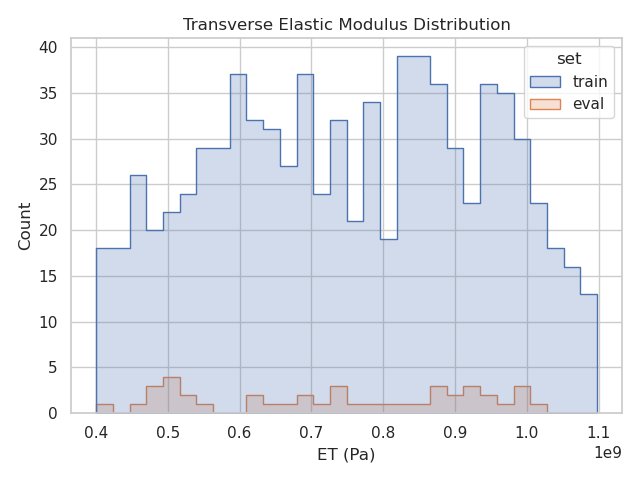}
        \caption{Seed 2}
    \end{subfigure}
    \begin{subfigure}[b]{0.45\textwidth}
        \includegraphics[width=\textwidth]{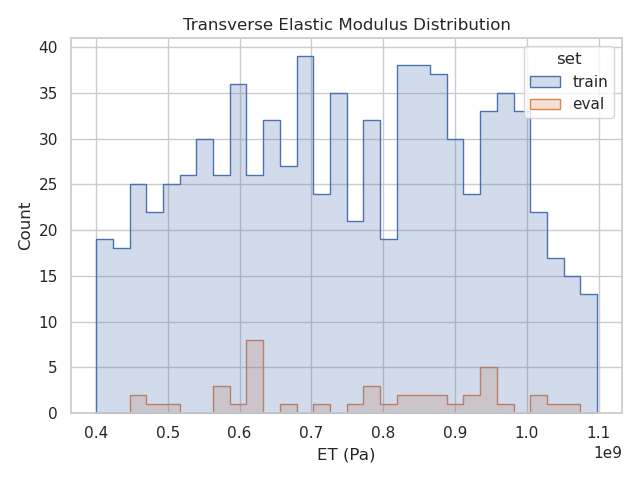}
        \caption{Seed 3}
    \end{subfigure}
    \caption{The $E_{\mathrm{T}}$ distribution of the training-evaluation split with different experiment seeds.}
    \label{fig:splits}
\end{figure}

The hyperparameters chosen for the models are summarized in Table~\ref{tab:hyperparams}. All models were optimized using the Adam optimizer~\cite{adam}. The Deformation U-Net was trained for 10~epochs with a batch size of~16 and a learning rate of~1$\times$10$^{-2}$, while for both the Stress U-Net and Modulus CNN, 50~epochs were required at a learning rate of~3$\times$10$^{-4}$. A train–validation split of 90–10 was used, where all steps generated from a given fiber simulation were kept within the same subset to prevent data leakage. The split was performed independently on the lumen and non-lumen samples and the resulting subsets were then concatenated, ensuring an even distribution of both fiber types across the training and validation sets. All experiments were conducted using three different random seeds to ensure model generalization. Figure~\ref{fig:splits} illustrates the distribution of the transverse Young’s modulus across the training and evaluation splits for different experimental seeds, showing consistent coverage of $E_{\mathrm{T}}$ range across all seeds. Lastly, the stride $\Delta$ was fixed at 4. Preliminary experiments showed that smaller values (e.g., $\Delta = 1$) resulted in negligible visual differences between consecutive frames, preventing the recursive U-Net from effectively learning the temporal evolution of the stress field. A stride of 4 introduces sufficiently pronounced changes to support stable learning, while still enabling prediction across five compression steps, thereby preserving an adequate level of temporal resolution.

All baseline models (Tables~\ref{tab:unet_metrics} and~\ref{tab:cnn_metrics}) were trained and evaluated under the same settings as the proposed models. Since the stress field baselines (StressNet, StressNet-mod, U-Net-mod, SCSNet) were originally designed for static, single-step prediction, their input layer was modified to accept the current stress field $S_i$, the current binary mask $B_i$, and the predicted next-step deformation mask $\hat{B}_{i+\Delta}$, matching the recursive formulation used by the proposed Stress U-Net. The regression baselines (ResNet-34, AlexNet, TI-CNN, CNN-custom) were adapted to take the final predicted stress field $S_{N-1}$ as a single-channel $256\times256$ input and output the scalar transverse Young's modulus prediction $\hat{E}_{\mathrm{T}}$.

To demonstrate the efficiency of the proposed deep learning pipeline, its computational requirements were compared against the FE method. While the initial simulation required a high-performance computing system (512~GB~RAM, 2.5~GHz~CPU) and averaged 5~minutes for a single fiber, the deep learning models were trained on a standard desktop NVIDIA~RTX~3060~GPU, taking only 60~seconds per epoch. Once trained, the models achieve a significant reduction in computational overhead, reducing inference time to under a second.

Accounting for training time, the total one-time cost of the deep learning pipeline is approximately 110 minutes (10 minutes for the Deformation U-Net, 50 minutes each for the Stress U-Net and Modulus CNN, at 60 seconds per epoch). Against this fixed cost, each avoided FE simulation saves approximately 5 minutes, while each inference adds a negligible fraction of a second. The break-even point, beyond which the trained surrogate becomes computationally advantageous over repeated FE simulation, is reached at approximately 22 fiber evaluations. Since the training dataset itself comprises 1,000 fibers, and any downstream application involving screening or design optimization would typically require evaluating many more candidate geometries, the framework is computationally favorable well within the scope of realistic use cases.

\captionsetup[subfigure][b!]{labelformat=empty}

\subsection{Evaluation Metrics}
For the quantitative evaluation of our proposed method, a set of complementary metrics was employed to assess both image sequence prediction and regression performance. All reported metrics represent the mean values across the different experimental seeds, with the corresponding standard deviations. 

For the image-to-image models, evaluation was performed using the Structural Similarity Index Measure (SSIM) \cite{SSIM} ranging from 0 to 1, where $1$ indicates perfect structural similarity, computed between the predicted and ground-truth images. In addition, the Root Mean Squared Error (RMSE) was used to quantify the pixel-wise deviation between the predicted and ground-truth stress fields. Because the stress fields are scaled by a single dataset-wide maximum, fibers with low peak stress occupy only a small portion of the [0, 1] range, and an unnormalized RMSE would be dominated by the few high-stress samples. To place all fibers on a comparable footing and isolate the fidelity of the predicted spatial distribution from differences in absolute magnitude, each field is rescaled to its own [0, 1] range before the error is computed. RMSE is thus defined as:
\begin{equation}
\mathrm{RMSE} =
\frac{1}{M}\sum_{j=1}^{M}
\sqrt{\sum_{i=0}^{N-\Delta} 
\big\| c^j \cdot \hat{S}_{i+\Delta}^{j} - c^j \cdot S_{i+\Delta}^{j} \big\|_2^2}
\end{equation}

where $c^j$ is the sample-specific scaling factor, defined by the ratio of the dataset's global maximum to the fiber's maximum stress,  which rescales each globally-scaled field to a per-sample [0, 1] range. The deformation masks are already binary and require no such rescaling. Lower RMSE values indicate better agreement between the predicted and ground-truth images.

For the regression model, the performance is evaluated using three complementary metrics. The coefficient of determination (R$^2$), ranging from 0 to 1, where 1 indicates a perfect fit, measures the proportion of variance in the target values explained by the model:

\begin{equation}
\mathrm{R}^2 = 1 - \frac{\sum_{j=1}^{M} \left( y^j - \hat{y}^j \right)^2}{\sum_{j=1}^{M} \left( y^j - \bar{y} \right)^2}
\end{equation}

where $y^j$ and $\hat{y}^j$ denote the ground-truth and predicted values, respectively, and $\bar{y} = \frac{1}{M}\sum_{j=1}^{M} y^j$ is the mean of the ground-truth values. The Mean Absolute Error (MAE), expressed in physical units, quantifies the average magnitude of prediction errors:

\begin{equation}
\mathrm{MAE} = \frac{1}{M}\sum_{j=1}^{M} \left| \hat{y}^j - y^j \right|
\end{equation}

The Mean Absolute Percentage Error (MAPE), expressed in percentage (\%), provides a scale-independent measure of relative error:

\begin{equation}
\mathrm{MAPE} = \frac{100}{M}\sum_{j=1}^{M} \left| \frac{\hat{y}^j - y^j}{y^j} \right|
\end{equation}

\subsection{Experimental Results}
The performance of the proposed framework is evaluated on both the image-to-image and regression components, compared against representative baseline models from the literature.

\begin{table}[b!]
\centering
\caption{Comparative evaluation of stress field prediction models}
\begin{tabular}{c c c c} \hline \textbf{Model} & \textbf{SSIM} (0--1) $\uparrow$ & \textbf{RMSE} (0--1) $\downarrow$ \\ \hline 
Stress U-Net (ours) & \textbf{0.9574}$\pm$\textbf{0.0074} & \textbf{0.0345}$\pm$\textbf{0.0079}\\
StressNet \cite{Nie_2019} & 0.9372$\pm$0.0048 & 0.0438$\pm$0.0022\\
StressNet-mod \cite{jcs7080311, jcs8100387} & 0.9172$\pm$0.0240 & 0.0479$\pm$0.0094\\
U-Net-mod \cite{BHADURI2022109879} & 0.9170$\pm$0.0218 & 0.0563$\pm$0.0090\\
SCSNet \cite{Nie_2019} & 0.8076$\pm$0.0290 & 0.0894$\pm$0.0023\\ \hline  
\end{tabular}
\label{tab:unet_metrics}
\end{table}

\textbf{Image-to-Image Model.} Table~\ref{tab:unet_metrics} reports the stress field prediction performance of the proposed Stress U-Net ($g$) alongside four baseline architectures drawn from the stress field prediction literature. The Stress U-Net achieves an SSIM of $0.9574 \pm 0.0074$ and an RMSE of $0.0345 \pm 0.0079$, with low variance across experimental seeds indicating robust generalization. The compared models include StressNet~\cite{Nie_2019}, an encoder--decoder network originally proposed for stress prediction in linear elastic materials. We also consider a StressNet variant used in~\cite{jcs7080311, jcs8100387}, referred to here as StressNet-mod, which omits the skip connections between encoder and decoder. In addition, we benchmark against U-Net-mod~\cite{BHADURI2022109879}, a slightly modified version of the standard U-Net applied to composite stress field prediction, and SCSNet~\cite{Nie_2019}, a lighter segmentation-based network from the same work. The proposed model outperforms all baseline models on both metrics, demonstrating that its design is well-suited to the dynamic nature of the task. The final configuration was selected through extensive hyperparameter tuning, and the contribution of individual architectural choices to this performance is examined in detail in the ablation study section (\ref{subsection:ablation}). In contrast, the baseline architectures were originally developed for idealized geometries or single-step stress prediction. In our dynamic setting, we operate on individual fibers with synthetic geometries derived from real plant-based fibers, which further increases the complexity of the task.

\begin{figure}[t!]
    \centering
    \includegraphics[width=\textwidth]{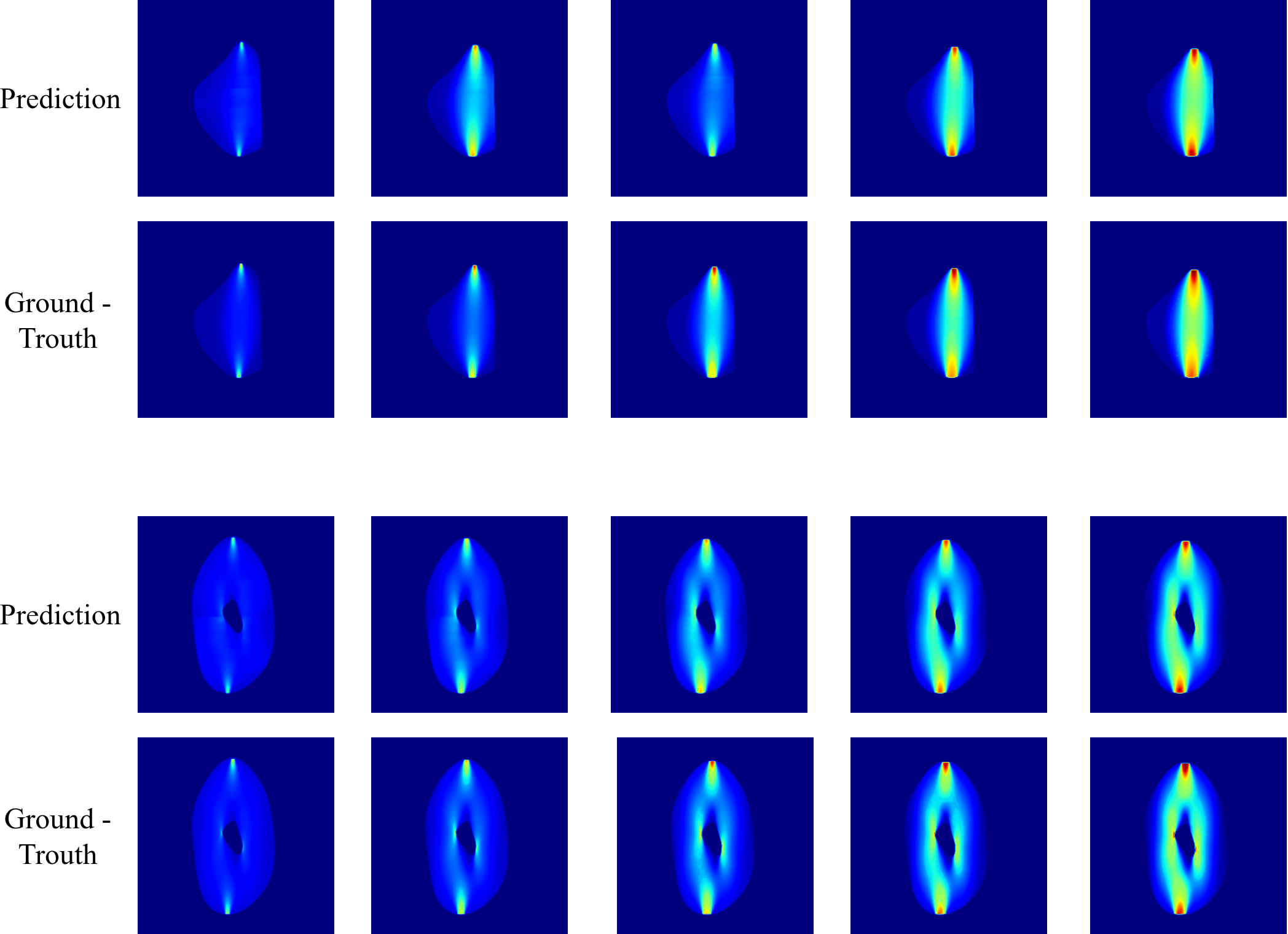}
\caption{Qualitative evaluation of the proposed framework for stress field prediction of individual synthetic plant fibers under transverse compression.}
\label{fig:examples}
\end{figure}

Figure~\ref{fig:examples} presents representative examples of the predicted versus ground-truth stress fields for individual synthetic plant fibers under transverse compression. Each pair of rows corresponds to a single fiber, with the top row showing the predicted stress field and the bottom row showing the ground-truth. Columns indicate sequential compression steps, highlighting the temporal evolution of the stress distribution. The first sample corresponds to a fiber without a lumen, while the second sample contains a lumen, illustrating the model’s ability to capture structural variations. For visualization purposes, the physical stress fields predicted by $g$ are mapped back to the JET colormap. Overall, the figure demonstrates that the models accurately reproduce both the spatial patterns and magnitudes of the stress field across different fiber geometries, illustrating the robustness of the framework.

\textbf{Regression Model.} Table~\ref{tab:cnn_metrics} reports the performance of the Modulus CNN ($h$) alongside four baseline models adapted to the same regression task. The proposed model achieves an R$^2$ of $0.938 \pm 0.024$, an MAE of $24.31 \pm 1.23$~MPa, and a MAPE of $5.25 \pm 0.91\%$, outperforming all baselines across every metric. The compared models include ResNet-34~\cite{he2015deepresiduallearningimage} and AlexNet~\cite{NIPS2012_c399862d}, well-established convolutional architectures that have shown strong performance across a broad range of tasks, including in the context of fiber composite analysis~\cite{jcs7080311}. Other architectures are the TI-CNN~\cite{Wang2021}, a temporal architecture originally proposed for predicting maximum internal stress during fracture in brittle materials, and a customised CNN~\cite{raaghav2025explainablepredictionmechanicalproperties}, referred to here as CNN-custom, developed for Young's modulus prediction from composite microstructure images. The proposed model outperforms all baselines across every metric. This is primarily attributed to its use of the final predicted stress field as input rather than the raw fiber geometry, allowing the learned representation of the full compression response to directly inform the modulus estimate, a formulation that none of the baselines exploit.

The regression model estimates the transverse Young's modulus from the final predicted stress field alone, rather than from the full predicted sequence. This design choice reflects the fact that the modulus is a converged mechanical property evaluated at the final compression state, and Jawad's inverse identification procedure (Eq. 1) is itself applied using force displacement data at the corresponding step. Nonetheless, incorporating the full stress sequence could plausibly improve regression accuracy by allowing the model to learn from the trajectory of stress accumulation rather than a single endpoint, for instance through a recurrent or attention based architecture applied across the five predicted frames. We consider this a promising direction, particularly for fibers where stress evolution is non monotonic due to buckling or self-contact effects.

\begin{table}[t!]
\centering
\caption{Comparative evaluation of regression models for transverse Young's modulus prediction}
\begin{tabular}{c c c c} \hline \textbf{Model} & \textbf{R$^2$} (0--1) $\uparrow$ & \textbf{MAE} (MPa) $\downarrow$ & \textbf{MAPE} (\%) $\downarrow$ \\ \hline
Modulus CNN (ours) & \textbf{0.938}$\pm$\textbf{0.024} & \textbf{24.31}$\pm$\textbf{1.23} & \textbf{5.25}$\pm$\textbf{0.91} \\ 
ResNet-34 \cite{he2015deepresiduallearningimage} & 0.895$\pm$0.059 & 30.48$\pm$6.84 & 6.89$\pm$1.70 \\ 
AlexNet \cite{NIPS2012_c399862d} & 0.881$\pm$0.088 & 28.42$\pm$3.91 & 6.67$\pm$0.49 \\ 
TI-CNN \cite{Wang2021} & 0.837$\pm$0.079 & 35.94$\pm$7.07 & 8.10$\pm$0.98 \\ 
CNN-custom \cite{raaghav2025explainablepredictionmechanicalproperties} & 0.754$\pm$0.014 & 48.04$\pm$8.89 & 10.49$\pm$2.11 \\ \hline
\end{tabular}
\label{tab:cnn_metrics}
\end{table}

\textbf{Performance at the Extremes of the $E_{\mathrm{T}}$ Distribution.} To assess whether the Modulus CNN exhibits systematic bias for fibers at the extremes of the transverse Young's modulus distribution shown in Fig.~\ref{fig:splits}, the held-out test predictions from the three evaluation seeds were pooled (70 samples in total) and split at the pooled median true $E_{\mathrm{T}}$ value into a low-modulus group ($E_{\mathrm{T}} \leq$ median) and a high-modulus group ($E_{\mathrm{T}} >$ median), using the same threshold across seeds. For each group, MAPE, MAE, R$^2$, and the mean signed percentage error (bias, $\frac{1}{n}\sum_i \frac{\hat{y}_i - y_i}{y_i}\times 100\%$) were computed per seed and averaged (Table~\ref{tab:et-extremes}).

\begin{table}[t!]
\centering
\caption{Prediction performance of the Modulus CNN, stratified by true modulus, averaged over 3 seeds ($\pm$ std across seeds). Groups are split at the pooled median true $E_{\mathrm{T}}$. $n$ is the number of held-out test samples per group, pooled over the 3 seeds.}
\begin{tabular}{l c c c c c} \hline
\textbf{Group} & \textbf{$n$} & \textbf{MAPE (\%)} & \textbf{Bias (\%)} & \textbf{MAE (MPa)} & \textbf{R$^2$} \\ \hline
Overall & 70 & $5.25 \pm 0.91$ & $-0.60 \pm 1.70$ & $24.31 \pm 1.23$ & $0.938 \pm 0.024$ \\
Low-modulus ($E_{\mathrm{T}} \leq$ median) & 35 & $5.52 \pm 2.14$ & $+0.15 \pm 2.26$ & $19.28 \pm 4.82$ & $0.917 \pm 0.033$ \\
High-modulus ($E_{\mathrm{T}} >$ median) & 35 & $5.04 \pm 0.36$ & $-1.66 \pm 2.01$ & $29.93 \pm 4.38$ & $0.595 \pm 0.297$ \\ \hline
\end{tabular}
\label{tab:et-extremes}
\end{table}

The relative error (MAPE) is comparable between the two groups ($5.52\%$ vs.\ $5.04\%$), indicating that the model does not lose proportional accuracy at either extreme of the distribution. The mean signed bias, however, reveals a mild systematic trend: high-modulus predictions are on average biased low by $1.66\%$, whereas low-modulus predictions show a negligible bias of $+0.15\%$. The larger absolute MAE for the high-modulus group ($29.93$ vs.\ $19.28$~MPa) is expected given the larger $E_{\mathrm{T}}$ values in this range, and is consistent with the modestly higher MAPE-normalized error observed there. The R$^2$ for the high-modulus group is markedly lower and more variable across seeds ($0.595 \pm 0.297$) than for the low-modulus group ($0.917 \pm 0.033$). Since R$^2$ is sensitive to the variance of the target within the evaluated subset, this is attributed primarily to the smaller within-group target range at this sample size ($\leq 14$ per seed), rather than to a genuine loss of accuracy, with MAE and MAPE remaining the more reliable indicators of absolute predictive accuracy for this subset. Overall, the model does not exhibit a large systematic bias at either extreme of the $E_{\mathrm{T}}$ distribution, though the small, consistent tendency to underestimate the highest-modulus fibers is a limitation to consider for engineering applications requiring accurate prediction of the stiffest fibers.

\textbf{Performance by Lumen Presence.} To quantify the model's robustness to the structural variation introduced by the possible presence of a lumen, the same held-out test predictions were instead split by lumen presence. Because the train--test split is randomized per seed rather than stratified by lumen presence, the two subsets are of unequal, seed-dependent size (46 and 24 pooled held-out samples for the no-lumen and with-lumen groups, respectively). Results are reported in Table~\ref{tab:et-lumen}.

\begin{table}[t!]
\centering
\caption{Prediction performance of the Modulus CNN, stratified by lumen presence, averaged over 3 seeds ($\pm$ std across seeds). $n$ is the number of held-out test samples per group, pooled over the 3 seeds.}
\begin{tabular}{l c c c c c} \hline
\textbf{Group} & \textbf{$n$} & \textbf{MAPE (\%)} & \textbf{Bias (\%)} & \textbf{MAE (MPa)} & \textbf{R$^2$} \\ \hline
Overall & 70 & $5.25 \pm 0.91$ & $-0.60 \pm 1.70$ & $24.31 \pm 1.23$ & $0.938 \pm 0.024$ \\
No lumen & 46 & $5.11 \pm 0.11$ & $-0.59 \pm 1.53$ & $28.09 \pm 2.40$ & $0.797 \pm 0.104$ \\
With lumen & 24 & $5.24 \pm 2.60$ & $-0.88 \pm 3.03$ & $16.48 \pm 6.61$ & $0.930 \pm 0.038$ \\ \hline
\end{tabular}
\label{tab:et-lumen}
\end{table}

The relative error (MAPE) is nearly identical between the two groups ($5.11\%$ without lumen vs.\ $5.24\%$ with lumen), and the mean signed bias is small and comparable in both cases ($-0.59\%$ vs.\ $-0.88\%$), indicating that the presence of a lumen does not introduce a systematic prediction bias. The lower absolute MAE for the with-lumen group ($16.48$ vs.\ $28.09$~MPa) reflects the narrower absolute $E_{\mathrm{T}}$ range spanned by that subset rather than higher accuracy per se, with the scale-normalized MAPE showing no meaningful gap between groups. As with the modulus-extremes analysis above, the lower R$^2$ observed for the no-lumen group ($0.797 \pm 0.104$ vs.\ $0.930 \pm 0.038$) is attributed to R$^2$'s sensitivity to within-group target variance at this sample size, rather than to a genuine accuracy gap, since MAPE and bias are essentially indistinguishable between the two groups. Overall, these results demonstrate that the fiber-specific stress concentrations introduced by the lumen do not translate into higher relative error or systematic bias for the $E_{\mathrm{T}}$ predictions.

\subsection{Ablation Study}
\label{subsection:ablation}

\begin{table}[b!]
\centering
\caption{Ablation study of the image-to-image model}
\begin{tabular}{c c c c} \hline \textbf{Model} & \textbf{Target} & \textbf{SSIM} (0--1) $\uparrow$ & \textbf{RMSE} (0--1) $\downarrow$ \\ \hline
Full & Deformation & \textbf{0.9822}$\pm$\textbf{0.0015} & \textbf{0.0761}$\pm$\textbf{0.0037} \\ 
Stride-1 & Deformation & $0.9805 \pm 0.0024$ & $0.0820 \pm 0.0053$ \\ \hline
Full & Stress & \textbf{0.9574}$\pm$\textbf{0.0074} & \textbf{0.0345}$\pm$\textbf{0.0079}\\
Stride-1 & Stress & $0.8923 \pm 0.0083$ & $0.0734 \pm 0.0102$ \\
No-def & Stress & $0.8235 \pm 0.0290$ & $0.0903 \pm 0.0137$ \\
No-rec & Stress & $0.5162 \pm 0.0316$ & $0.1333 \pm 0.0014$ \\ \hline 
\end{tabular}
\label{tab:ablation}
\end{table}

To assess the contribution of each design decision in the proposed framework, three ablation variants of the image-to-image model are evaluated and compared against the full model in Table~\ref{tab:ablation}.

\textbf{Stride-1} This variant applies the full two-U-Net recursive architecture, but uses a stride of $\Delta = 1$ instead of the adopted $\Delta = 4$. Both $f$ and $g$ are trained and evaluated with unit-step transitions, so that $\hat{B}_{i+1} = f_\theta(B_i)$ and $\hat{S}_{i+1} = g_\phi(S_i, B_i, \hat{B}_{i+1})$, requiring 20 recursive rollout steps to reach the same final compression state.

\textbf{No-def} This variant replaces the two-U-Net cascade with a single stress U-Net $g'_\phi$. Since the predicted deformation mask $\hat{B}_{i+\Delta}$ from $f$ is no longer available, $g'_\phi$ receives only two input channels: the initial binary mask $B_0$ and the current stress field $S_i$. The model is still trained and evaluated recursively with stride $\Delta = 4$, so that $\hat{S}_{i+\Delta} = g'_\phi(S_i, B_0)$ at each step.

\textbf{No-rec} This variant removes both the deformation U-Net and the recursive rollout mechanism entirely. A single U-Net $g''_\phi$ is trained in a one-shot, non-recursive fashion: it receives the initial binary mask $B_0$ as its sole input and directly outputs the full predicted stress sequence as a five-channel image, $g''_\phi(B_0) = (\hat{S}_\Delta, \hat{S}_{2\Delta}, \hat{S}_{3\Delta}, \hat{S}_{4\Delta}, \hat{S}_{5\Delta})$.

\begin{figure}[t!]
    \centering
    \includegraphics[width=\textwidth]{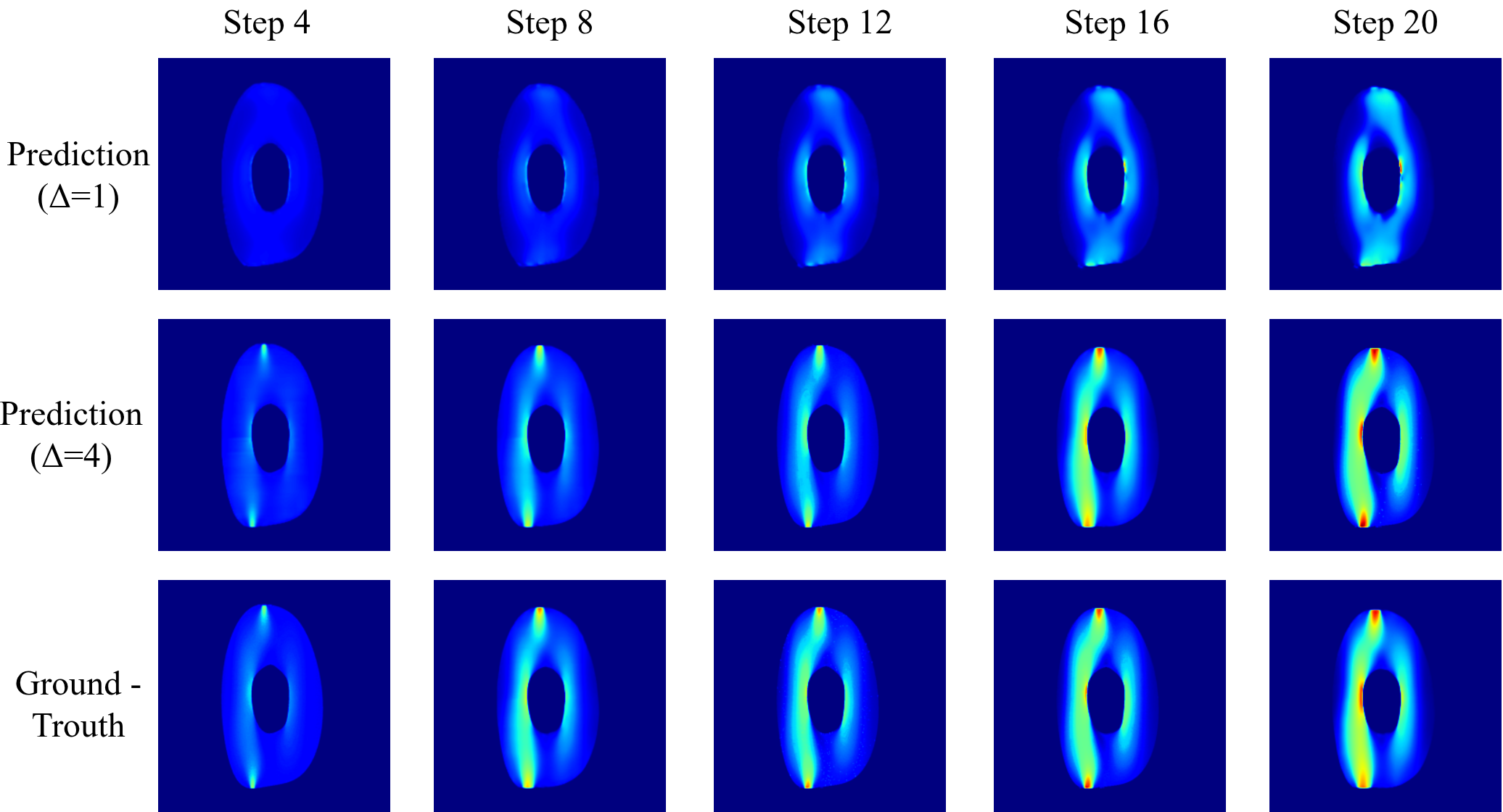}

\caption{Qualitative comparison of rollout predictions for stride values $\Delta=1$ and $\Delta=4$ across sequential evaluation steps.}
\label{fig:stride_comparison}
\end{figure}

The results in Table~\ref{tab:ablation} demonstrate that each design choice contributes meaningfully to the model's performance. For the deformation target, Stride-1 and the Full model yield nearly equivalent SSIM scores ($0.9805$ vs.\ $0.9822$). Because $B_i$ is a binary mask, even a unit-stride transition produces a discrete, non-negligible change at the fiber boundary, i.e., pixels either belong to the fiber or they do not, providing a learnable signal regardless of $\Delta$. The picture is markedly different for the stress target. The Full model achieves an SSIM of $0.9574$ and an RMSE of $0.0345$, whereas Stride-1 drops to $0.8923$ and $0.0734$, respectively, a substantial degradation despite its finer temporal resolution. This is further illustrated in Figure~\ref{fig:stride_comparison}, which shows that $\Delta=1$ rollout predictions exhibit noticeable loss of the stress flow compared to both the $\Delta=4$ predictions and the ground truth, consistent with the observation that unit-stride transitions produce near-identical consecutive frames, offering insufficient signal for the stress U-Net to learn meaningful temporal dynamics. Removing the deformation branch (No-def) leads to a further decline in stress prediction accuracy (SSIM~$=0.8235$, RMSE~$=0.0903$) and notably increased variance across seeds, highlighting the importance of the predicted future mask $\hat{B}_{i+\Delta}$ as an explicit geometric prior that guides the stress evolution at each step. The most severe degradation is observed for No-rec (SSIM~$=0.5162$, RMSE~$=0.1333$), which collapses to near-chance structural fidelity, underscoring that the recursive formulation is critical. 

In practice, this means intermediate steps between the five predicted frames (i.e., steps not divisible by $\Delta = 4$ are not directly produced by the model. If intermediate-step estimates are required, the most straightforward approach is linear interpolation between the two nearest predicted frames, since the underlying deformation process is smooth and monotonic in this small-strain regime. Because stress evolves nonlinearly near contact boundaries, such interpolation would introduce some additional error relative to a direct model prediction at that step, though this error is expected to be modest given the limited displacement range (1 µm over 20 steps) and the high fidelity already achieved at the five predicted anchor points (SSIM = 0.9574).

This severe degradation can be understood mechanistically. In the No-rec configuration, the single U-Net must map the undeformed fiber mask directly to five stress fields at once, effectively learning a static function from initial geometry to a five-step temporal sequence. This requires the network to implicitly encode the entire loading history within a single forward pass, without access to any intermediate mechanical state. Because the relationship between initial fiber geometry and the stress distribution at later compression steps is highly path-dependent, particularly near contact boundaries where local stress concentrations evolve nonlinearly with platen displacement, a purely feed-forward mapping lacks the inductive bias needed to represent this dependency. The recursive formulation instead decomposes the problem into a sequence of local, physically grounded updates, where each step's prediction is conditioned on the actual predicted state of the fiber and stress field at the previous step. This mirrors the incremental nature of the underlying finite element solution process itself, in which each loading step is resolved conditional on the deformed configuration from the previous step, and appears to be substantially easier for the network to learn than the equivalent one-shot mapping.

\section{Conclusion}\label{sec5}

In this study, a deep learning framework is proposed for the fast and accurate prediction of stress fields and mechanical properties in individual synthetic plant fibers under transverse compression. The framework consists of three specialized components: a Deformation U-Net that predicts the evolving fiber geometry step by step, a Stress U-Net that recursively predicts the physical stress field conditioned on the current stress state and the predicted deformation, and a Modulus CNN that estimates the transverse Young's modulus from the final predicted stress field. The framework recovers the full stress field sequence with high fidelity, achieving an SSIM of $0.9574 \pm 0.0074$ and an RMSE of $0.0345 \pm 0.0079$, and predicts the transverse Young's modulus with an R$^2$ of $0.938 \pm 0.024$ and a MAPE of $5.25 \pm 0.91\%$, outperforming all evaluated baseline architectures on every metric. These results are consistent across fibers with and without a lumen and across independent random seeds, demonstrating robust generalization of the proposed approach. An ablation study further confirms that the recursive formulation, the deformation branch, and the choice of stride are each critical to this performance.
Beyond predictive accuracy, an important advantage of the proposed framework is its computational efficiency. Whereas a single FE-based transverse compression analysis requires several minutes on a high-performance computing cluster, the trained models reduce inference time to under a second per fiber, enabling rapid characterization across large populations of fiber geometries without significant loss of accuracy, making the framework a practical surrogate for finite element simulation in bio-based composite design. A primary limitation of this work is that the framework is trained and evaluated entirely on synthetically generated fibers with FE-derived ground truth, not yet validated against physical single-fiber compression measurements. The underlying simulations also assume linear elastic, transversely isotropic material behavior, and the framework is currently restricted to transverse compression of individual fibers rather than full composite microstructures involving fiber-matrix and fiber-fiber interactions.
Future work will focus on experimental validation against physical measurements, extension to nonlinear or viscoelastic material behavior and alternative loading modes, domain adaptation to fiber types with different elastic properties, and incorporation of uncertainty quantification to support engineering design decisions.

\section*{Declarations}

\bmhead{Funding}
The work was supported by the European Union through the RAIDO Project under Grant No. 101135800.

\bmhead{Acknowledgements}
Support from the EIPHI Graduate School (ANR-17-EURE-0002) is also acknowledged.

\bmhead{Conflict of interest}
All authors certify that they have no affiliations with or involvement in any organization or entity with any financial interest or non-financial interest in the subject matter or materials discussed in this manuscript.

\bibliography{sn-bibliography}

\end{document}